\documentclass{article} 

\PassOptionsToPackage{hyphens}{url}

\usepackage{iclr2023_conference,times}
\usepackage{CJK} 
\usepackage{pdflscape}
\usepackage{graphicx}
\usepackage{tikz}

\usepackage{amsmath,amsfonts,bm}

\def\eqref#1{equation~\ref{#1}}

\def\1{\bm{1}}

\DeclareMathAlphabet{\mathsfit}{\encodingdefault}{\sfdefault}{m}{sl}
\SetMathAlphabet{\mathsfit}{bold}{\encodingdefault}{\sfdefault}{bx}{n}

\usepackage{rotating}
\usepackage{microtype}
\usepackage{makecell}
\usepackage{amsmath}
\usepackage{booktabs}
\usepackage{colortbl}
\usepackage[utf8]{inputenc}
\definecolor{lightgray}{rgb}{0.9,0.9,0.9}
\usepackage{caption}
\usepackage{subcaption}
\usepackage{xcolor}
\usepackage{graphicx}
\usepackage{setspace}
\usepackage[hidelinks]{hyperref}
\usepackage{url}
\usepackage{multirow}
\usepackage{colortbl}
\usepackage{tabularx}
\usepackage{blindtext}
\usepackage{pgfplots}
\pgfplotsset{compat=1.18} 
\usepackage{tikz}
\usetikzlibrary{er,positioning,bayesnet}
\usepackage[inline]{enumitem}
\usepackage{makecell}
\usepackage{siunitx}
\usepackage{nicefrac}
\usepackage{listings}
\usepackage[raster,skins]{tcolorbox} %
\usepackage{adjustbox}
\usepackage{cleveref}

\title{Baichuan4-Finance Technical Report}
\author{
\\
\parbox{\linewidth}{Hanyu Zhang\thanks{These authors contributed equally.}, \hspace{1em}Boyu Qiu\footnotemark[1], \hspace{1em}Yuhao Feng,
\hspace{1em}Shuqi Li, \\
\hspace{1em}Qian Ma, 
\hspace{1em}Xiyuan Zhang\thanks{Corresponding Author.}, \hspace{1em}Qiang Ju, \hspace{1em}Dong Yan, \hspace{1em}Jian Xie\footnotemark[2]} 
\AND
Baichuan Inc.\\
}

\iclrfinalcopy
\begin{document}
\begin{CJK}{UTF8}{gbsn}

\maketitle

\begin{abstract}

Large language models (LLMs) have demonstrated strong capabilities in language understanding, generation, and reasoning, yet their potential in finance remains underexplored due to the complexity and specialization of financial knowledge. In this work, we report the development of the Baichuan4-Finance series, including a comprehensive suite of foundational \textbf{Baichuan4-Finance-Base} and an aligned language model \textbf{Baichuan4-Finance}, which are built upon Baichuan4-Turbo base model and tailored for finance domain. Firstly, we have dedicated significant effort to building a detailed pipeline for improving data quality. Moreover, in the continual pre-training phase, we propose a novel \textbf{domain self-constraint} training strategy, which enables Baichuan4-Finance-Base to acquire financial knowledge without losing general capabilities. After Supervised Fine-tuning and Reinforcement Learning from Human Feedback and AI Feedback, the chat model Baichuan4-Finance is able to tackle various financial certification questions and real-world scenario applications. We evaluate Baichuan4-Finance on many widely used general datasets and two holistic financial benchmarks. The evaluation results show that Baichuan4-Finance-Base surpasses almost all competitive baselines on financial tasks by significant margins without sacrificing performance on general LLM benchmarks. At the same time, Baichuan4-Finance demonstrates even more impressive performance on financial application scenarios, showcasing its potential to foster community innovation in the financial LLM field. 

\end{abstract}
\clearpage

\tableofcontents
\clearpage

\section{Introduction}
\label{sec:intro}

Since the launch of ChatGPT~\footnote{\url{https://openai.com/index/chatgpt/}}, interest in large language models (LLMs) has surged worldwide. The release of the Llama series~\citep{touvron2023llama} further sparked excitement within the open-source community, particularly for GPT-level local LLMs. Recently, Claude-3 Opus~\footnote{\url{https://www-cdn.anthropic.com/de8ba9b01c9ab7cbabf5c33b80b7bbc618857627/Model\_Card\_Claude\_3.pdf}}, Gemini-1.5~\citep{team2023gemini}, GPT-4~\citep{achiam2023gpt}, GPT-4o (omni)~\footnote{\url{https://openai.com/index/hello-gpt-4o/}}, o1-preview and o1-mini~\footnote{\url{https://openai.com/index/learning-to-reason-with-llms/}}, the updated versions of ChatGPT, quickly rose to the top of the Chatbot Arena~\citep{chiang2024chatbot}. Additionally, Llama-3~\footnote{\url{https://github.com/meta-llama/llama3/blob/main/MODEL_CARD.md}} has become the leading open-weight model series, closing the performance gap with top proprietary models and regarded as on par with GPT-4. An increasing number of competitive LLMs, such as Baichuan~\citep{yang2023baichuan}, Qwen~\citep{bai2023qwen}, Mistral~\citep{jiang2023mistral}, and Gemma~\citep{team2024gemma}, are reported gradually, following in the footsteps of the GPT series and Llama series.

Applying general LLM to the finance domain presents challenges, as financial documents often contain complex numerical data and domain-specific terminology, requiring advanced numerical processing and reasoning skills. As a result, financial LLMs must have extensive domain knowledge to interpret the subtle implications. Recently, many financial LLMs have also been developed to meet the needs of the financial domain, such as ~\citep{wu2023bloomberggpt,yang2023fingpt, liu2021finbert, xie2023pixiu,yang2023investlm,shah2022flue}, which have demonstrated superior capabilities over general models in many financial-related tasks. These advancements have revealed the potential of unstructured data for data-driven financial decision-making and for transforming financial documents into actionable insights and market intelligence. 

Over the past two years, Baichuan Inc. has introduced many LLM series, including Baichuan series~\footnote{\url{https://huggingface.co/baichuan-inc/Baichuan-7B, https://huggingface.co/baichuan-inc/Baichuan-13B-Chat}}, Baichuan2 series~\citep{yang2023baichuan}, Baichuan3 series and progressed to the latest Baichuan4 series~\footnote{\url{https://platform.baichuan-ai.com/playground?initialmodel=Baichuan4}}. During the same time, we reported the role play LLMs Baichuan-NPC series~\footnote{\url{https://npc.baichuan-ai.com/index}} providing highly flexible personalized character customization capabilities. In this work, we introduce the first financial LLM series of the Baichuan LLM family: Baichuan4-Finance. It is a series of LLMs grounded in the Baichuan4-Turbo base model~\footnote{\url{https://platform.baichuan-ai.com/playground?initialmodel=Baichuan4-Turbo}} and specifically tailored for the finance domain. This LLM series includes a foundational/base language model Baichuan4-Finance-Base, which is pre-trained but not yet aligned with human preferences, as well as an instruction-tuned model Baichuan4-Finance, which has been fine-tuned for chat and downstream applications.

Generally, the development of a domain-specific LLM consists of two main stages: 

\begin{itemize}
    \item \textbf{Continual pre-training}: aiming to remain an existing general LLM with new domain-specific data on an incremental sequence of tasks~\citep{lee2024survey}.
    \item \textbf{Alignment}: aiming to fine-tune models to follow instructions, align with human preferences, and improve LLMs' specific domain capabilities~\citep{dubey2024llama}. 
\end{itemize}

Based on past experience, many factors have a significant impact on the performance of the final model: data (quality \& size \& mixture ratio), model (tokenizer \& architecture), and training strategy (continual pre-training \& alignment). In this report, we will introduce how we seek to optimize the above factors throughout our development process.

\section{Tokenizer \& Model Architecture}

\label{sec:arch}

This section introduces the tokenizer and model architecture of Baichuan4-Finance-Base.

\subsection{Tokenizer}

We employ the byte-level byte-pair encoding (BBPE)~\citep{wang2020neural} as the tokenizer, which has been proven to have a high compression rate relative to others~\citep{yang2024qwen2}. The resulting vocabulary has 141,056 regular tokens. 

\subsection{Model Architecture}

\paragraph{Pre-normalization with RMSNorm}

To handle training stability, we employ the RMSNorm~\citep{jiang2024pre} for pre-normalizing.

\paragraph{Grouped-Query Attention}

We leverage the Grouped Query Attention (GQA)~\citep{ainslie2023gqa} instead of the traditional Multi-Head Attention (MHA)~\citep{vaswani2017attention} to improve inference speed and to reduce the size of key-value caches during decoding. 

\paragraph{Positional Encoding}

RoPE~\citep{su2023enhanced} is employed to serve for positional encoding as most open-sourced models do.

\section{Continual pre-training}

\label{sec:pre}

In the pre-training of the Baichuan4-Finance-Base, we mainly focused on dataset composition, dataset quality enhancement, determining the data mixture ratio with scaling law, and the strategy of continuing pre-training. We present each of these components separately below.

\subsection{Pre-training Data Composition}

The Pre-training Data consists of two main parts: general data encapsulating world knowledge and financial data containing extensive financial knowledge.

\paragraph{General Data (400B tokens - 80\% of pre-training data)}

To create a comprehensive global knowledge system, we collect diverse data from various sources, such as internet webpages, books, research papers, codebases, and more.

\paragraph{Financial Data (100B tokens - 20\% of pre-training data)}

To teach the model finance-related knowledge, we construct a comprehensive dataset comprising a range of financial documents including news, press releases, finance books, finance journals, and social media posts. The composition of the financial corpus is shown in Figure~\ref{fig:pretrain_data}.

\subsection{Data Quality Enhancement}

Generally, datasets that are unfiltered or only lightly filtered tend to have lower quality compared to more carefully curated ones. Therefore, we have carefully designed a pipeline to enhance the overall quality of the data. We filter and clean the associated texts using rule-based heuristics, such as controlling the average sentence length and document length. Then the following processing methods are employed to the dataset step by step.

\paragraph{Data Filtering with Quality Classifier}

To improve data quality, we first train an automatic quality classifier based on logistic regression to remove the low-quality documents. High-quality examples are selected from general data, while the raw unfiltered documents are sampled from various sources and taken as low-quality examples. We mix these two class data with a 1:1 ratio for classifier training. Once the classifier is fitted, we then apply it to score the unfiltered documents and re-sample them by giving priority to those predicted to be of higher quality:

\begin{equation}
    \mathtt{np.random.pareto}(\alpha) > 1 - \mathtt{document\_score}.
    \label{eq:pareto}
\end{equation}

The meaning of this filtering rule is that a document is retained only if a random sample from the Pareto distribution is greater than $1 - \texttt{document\_score}$. This setup results in a higher retention probability for documents with higher quality scores.

\paragraph{Data Filtering with Model Scoring}

After the quality classifier, we perform further data filtering from a more fine-grained perspective. At first, by prompting Baichuan4-Turbo, we score millions of data samples across multiple dimensions, including readability, coherence, informativeness, safety, degree of anonymity, and unattainable references. The scoring system is designed to capture both qualitative and quantitative aspects of the data reliably and responsibly. Secondly, the data and corresponding labels serve as inputs to train an automatic scoring model grounded on XLM-Roberta~\footnote{\url{https://huggingface.co/docs/transformers/model\_doc/xlm-roberta}}. Lastly, after the scoring model is fitted, we leverage the aforementioned $\mathtt{pareto}$ re-sampling strategy to filter data.

In this process, we also leverage an abnormal loss detection schema to filter low-quality data. Specifically, we obtain the loss of every source of data using a 1B Base model and perform sampling analysis on data with abnormal loss values. If the anomalies are due to low-quality data and represent common, generalizable issues, we apply rules to identify and remove similar entries. For more complex issues that cannot be fully addressed with rules, we collect a sample set and feed them into the XLM-RoBERTa model to learn their patterns.

\paragraph{Code, Math and Markdown Data}

Due to the significantly different token distributions of code, math, and markdown data compared to natural language, we develop specialized filtering models tailored to these data types.

\paragraph{Data De-duplication \& Anonymization}

To prevent redundancy, following~\citep{dubey2024llama}, we perform many rounds of de-duplication processes at the URL level, document level and line level, respectively.

\begin{itemize}
    \item URL level de-duplication: we retain the latest version of each page corresponding to its URL across the entire dataset.
    \item Document level de-duplication: we leverage global MinHash~\citep{broder1997resemblance} across the entire dataset to de-duplicate the similar docments. Remarkably, different de-duplication thresholds are set for documents of varying difficulty. 
    \item Line-level de-duplication: following Llama 3~\citep{dubey2024llama} and ccNet~\citep{wenzek2019ccnet}, we remove lines that occur more than certain times within each bucket.
\end{itemize}

Besides, We use regular expressions to anonymize data, including personal names, paper references, etc.

\subsection{Determining the Data Mixture Ratio by Scaling Laws}

\paragraph{Scaling Laws} 

Considering the differences in data distribution between the pre-training and continual pre-training stages, careful attention must be given to seeking the optimal mixture ratio of the pre-training financial data to develop a high-quality financial LLM. To obtain the Baichuan4-Finance-Base, we need to determine the mixture ratio of $n=37$ financial data sources, including research reports, academic papers, examination questions, and more.   

To identify the optimal data mixture ratio with limited training costs, we conduct a two-stage scaling law methodology by training several small models on specific data mixture ratios and using their performance to estimate how a larger model would perform with the same mixture ratio:

\begin{itemize}
    \item We first establish a correlation between the model size, data mixture ratio, dataset sizes and the model validation loss with D-CPT Law~\citep{que2024d}.
    Assuming that $N$ indicates the model size, $\bm{D}=[D_1, D_2, \cdots, D_n]$ indicates the dataset size of $n$ data sources, given a specific mixture ratio $\bm{r}=[r_1, r_2, \cdots, r_n]$ corresponding to these data sources, we could leverage the parameterization function $L=\{L_i\}_{i=1}^n$ defined as follows to predict the validation loss for each data source:
    \begin{equation}
    L=L_i(N, D_i, r_i)=E_i+\frac{A_i}{N^{\alpha_i}}+\frac{B_i \cdot r_i^{\eta_i}}{D_i^{\beta_i}}+\frac{C_i}{r_i^{\prime \gamma}}, \text { where } r_i^{\prime}=r_i+\epsilon.
    \end{equation}
    $\epsilon$ is used to handle the scenarios when $r_i$ approaches 0, and $\{E_i, A_i, B_i, C_i, \alpha_i, \beta_i, \gamma_i, \eta_i\}_{i=1}^n$ are parameters need to learn. 
    
    \item Then, we correlate the validation loss with the performance of pre-training benchmark tasks, i.e. to predict the accuracy on benchmark datasets ~\citep{dubey2024llama}. Specifically, we leverage a sigmoid function to model the relation between validation losses and benchmark accuracies for all domains.

\end{itemize}

With these two scaling laws, we are able to predict the downstream performance of arbitrary data mixture ratios, model size, and dataset sizes.

\paragraph{Scaling Law Results}

\begin{figure}
 \centering
\includegraphics[width=12cm]{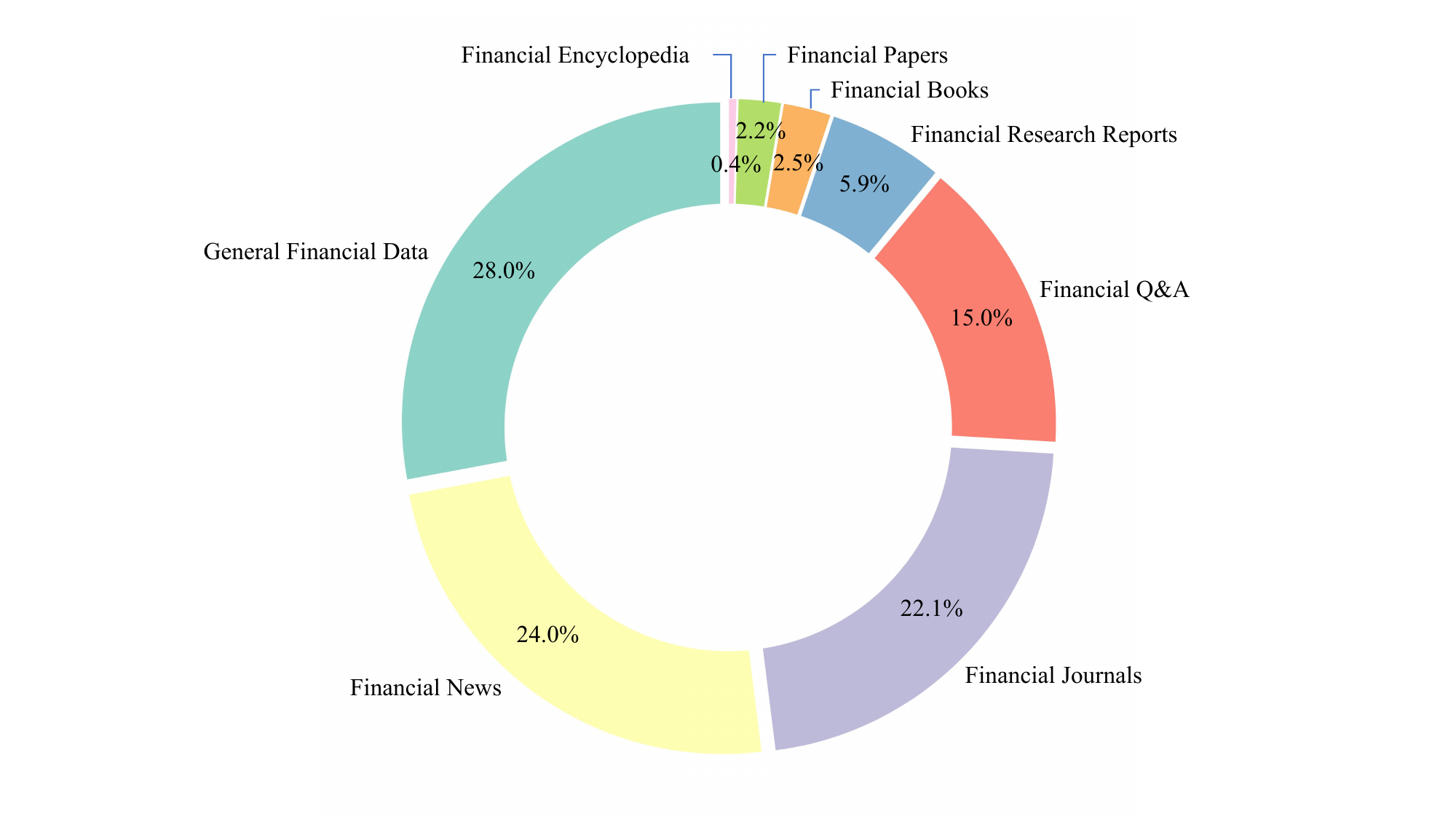}
\caption{The composition of financial data for Baichuan4-Finance-Base pre-training.}
\label{fig:pretrain_data}
\end{figure}

Based on the scaling laws, we could determine the financial pre-training data mixture ratio, which is visualized in Figure~\ref{fig:pretrain_data}. 

\subsection{Continual pre-training}

We continue training Baichuan4-Turbo base model to develop Baichuan4-Finance-Base. Regular continual pre-training involves training on new data using the same training strategy as the pre-training phase. Drawing inspiration from the concept of PPO~\citep{schulman2017proximal}, we introduce a novel domain-specific continual pre-training strategy, termed \textbf{domain self-constraint continual pre-training}. This approach balances two objectives: firstly, maintaining the knowledge of the general model, ensuring the retention of general task-solving capabilities; and secondly, enabling the training model to acquire domain-specific financial knowledge. To achieve these two goals, we design two different objectives, respectively.

Considering a reference LLM - Baichuan4-Turbo base model parameterized by $\theta_{ref}$, and the Baichuan4-Finance-Base parameterized by $\theta_{fin}$ and a pre-training document $\bm{x}= \{x_t\}^T_{t=1}$, where $T$ indicates document length.

If the document $\bm{x}$ belongs to the general data, we formalize the learning objective as:

\begin{equation}
\mathcal{L} = \mathcal{L}_{\mathrm{kl}}=\frac{1}{T} \sum_{t=1}^T KL\left(P_{\theta_{fin}}\left(x_t \mid \boldsymbol{x}_{<t}\right), P_{\theta_{ref}}\left(x_t \mid \boldsymbol{x}_{<t}\right)\right),
\end{equation}
which enables Baichuan4-Finance-Base to maintain the general knowledge. To improve the complexity efficiency, we sample 200 probabilities of the reference model for each token.

While the document $\bm{x}$ belongs to the financial data, the objective consists of two parts formalized as follows:

\begin{equation}
\mathcal{L} = \alpha\mathcal{L}_{\mathrm{kl}} + \mathcal{L}_{\mathrm{lm}},
\end{equation}
in which $\alpha$ is the hyper-parameter to balance the two parts of this objective, and $\mathcal{L}_{\mathrm{lm}}$ is defined as follows,
\begin{equation}
\mathcal{L}_{\operatorname{lm}}=\frac{1}{T} \sum_{t=1}^T-\log P_{\theta_{fin}}\left(x_t \mid \boldsymbol{x}_{<t}\right).
\end{equation}

With this objective, Baichuan4-Finance-Base learns financial domain knowledge while maintaining general knowledge.

By minimizing the above objective on the whole pre-training data,
\begin{equation}
\theta_{fin}^*=\arg \min _{\theta_{fin}} \mathcal{L},
\end{equation}
we could obtain the pre-trained financial model - Baichuan4-Finance-Base.

\subsection{Annealing}

During the continual pre-training stage on the final 80M tokens, the learning rate was linearly reduced to 0. Additionally, the data mixture ratio was adjusted to prioritize high-quality financial data sources ~\citep{dubey2024llama}.

\section{Alignment}

\label{sec:alignment}
 After the extensive large-scale pre-training phase, we further delve into the alignment phase to refine its capabilities across diverse domains such as financial computation, financial logical reasoning, and financial instruction following. This phase is crucial for ensuring that the model's outputs align with human values, making them useful, truthful, and safe. Specifically, we focus on collecting high-quality demonstration data for alignment training aiming to reduce the reliance on human labeling while maximizing data quality and reliability.

\subsection{Supervised Fine-tuning}

\subsubsection{Demonstration Data Construction}

The construction process of demonstration data mainly consists of three steps: finance materials construction, seed sample repository construction and large-scale demonstration construction.

\paragraph{Finance Materials Construction}

Firstly, we collect a large-scale financial corpus, including news articles, research papers, books, and other relevant financial documents, to serve as the foundational demonstration data source. 

\paragraph{Seed Sample Repository Construction}

Secondly, we construct a high-quality human-annotated seed sample repository for further large-scale automated data construction. From an application perspective, we categorize the problems that Baichuan4-Finance aims to solve along multiple dimensions to construct the instructions:
    \begin{itemize}
        \item By basic capabilities: classification, clustering, generation, summarization, extraction, question-answering, coding, mathematics, etc.
        \item By problem complexity: simple instructions, normal instructions, complex instructions.
    \end{itemize}
    For these instructions, we manually constructed high-quality data to serve as seeds for generating large-scale demonstration datasets.

\paragraph{Large-scale demonstration construction}

For one document sampled from the constructed finance materials repository, with an in-context learning strategy, in which the examples are sampled from the seed sample repository, we prompt Baichuan4-Turbo to generate the corresponding instruction. Then, multiple responses to an instruction are obtained using diverse generation strategies. Lastly, human annotators manually review and filter the generated demonstration data and construct a high-quality demonstration dataset for further supervised fine-tuning.

\subsubsection{Training Strategy}
We have curated a comprehensive instruction dataset containing over 160,000 high-quality demonstrations, covering tasks such as instruction following, financial classification, clustering, and generation.  Based on this dataset and the regular supervised fine-tuning process, the supervised fine-tuned model is produced.

\subsection{Reward Model}

\subsubsection{Preference Data Construction}

\paragraph{Human Feedback Preference Data Construction}

For scenarios with non-unique answers, such as in information understanding and creative tasks, we use human feedback ~\citep{ouyang2022training} to construct the preference dataset. Specifically, we first collect prompts and perform multiple high-temperature samplings for each prompt using Baichuan4-Turbo. Then, human annotators score and rank each answer on truthfulness, harmlessness, fluency, and instruction following. Finally, based on the ranking results, we could construct many chosen and rejected pairs for each prompt.

\paragraph{AI Feedback Preference Data Construction}

For scenarios with unique answers, such as mathematical reasoning, which is particularly important for financial scenarios requiring calculations, we use AI feedback ~\citep{cui2024ultrafeedback,li2024boosting} to construct the preference dataset. For example, given a mathematical dataset, for each math prompt, Baichuan4-Turbo generates answers through several inference runs. A verifier then checks each answer against the ground truth label to determine its correctness. Samples where all answers are correct and those with all incorrect answers are removed. The remaining samples are used to construct the preference dataset.

By the above two processes, we fulfill the preference dataset construction, which is used for the reward model training.

\subsubsection{Reward Model Training}

The reward model is initialized from the Baichuan4-Turbo base model by replacing the language modeling head with a value head. For a prompt $x$ with its corresponding preference data $(y_{\text{chosen}}, y_{\text{rejected}})$, following ~\citep{ouyang2022training}, the loss function of the reward model is formulated as:

\begin{equation}
    \mathcal{L}_{rm} = -\frac{1}{N} \mathbb{E}_{(x, y_{\text{chosen}}, y_{\text{rejected}})} \left[ \log \left( \sigma \left( r_\theta(x, y_{\text{chosen}}) - r_\theta(x, y_{\text{rejected}}) \right) \right) \right],
\end{equation}
where $N$ denotes the number of preference pairs of the prompt $x$, $\theta$ denotes the parameter set of reward model and $r_\theta(x, y)$ indicates the output reward score.

\subsection{Reinforcement Learning from Human Feedback and AI Feedback}

Reinforcement Learning from Human Feedback (RLHF) or from AI feedback is increasingly regarded as a crucial approach for LLM alignment. However, alignment methods that depend on the reward model present notable challenges due to the inherent instability and imperfections of reward models, leading to problems like reward hacking and misalignment with human intentions. Inspired by ~\citep{yan2024reward}, we leverage such a PPO-based reward-robust framework to pursue more reliable and resilient reinforcement learning. We first initialize the model with the supervised fine-tuned Baichuan4-Finance-Base and use it to generate predictions for randomly selected prompts from the overall prompts constructed in preference data construction. Then we optimize it with the PPO strategy and the above reward-robust framework by maximizing the overall reward. After reinforcement learning, the final chat version model is produced, called Baichuan4-Finance.

\section{Experiments}
\label{sec:experiment}

\subsection{Benchmarks}

\paragraph{Public General Benchmarks}
 C-Eval~\citep{huang2024c}, CMMLU~\citep{hendrycks2020measuring}, MMLU~\citep{hendrycks2020measuring}, GSM8k\_ZH (translated from GSM8k~\citep{cobbe2021training} into Chinese), and HumanEval~\citep{chen2021evaluating}. Besides, we also leverage 7 math datasets from ~\citep{li2024numinamath}, including MATH, K-12, Orca-math, AoPS Forum, Olympiads, AMC\&AIME and Synthetic. We randomly sampled 800 samples from them and manually translated 400 of them into Chinese for testing.

\paragraph{Public Financial Benchmarks}
We evaluate the financial capabilities by using two benchmarks FinanceIQ and FLAME.

FinanceIQ~\footnote{\url{https://huggingface.co/datasets/Duxiaoman-DI/FinanceIQ}} is a Chinese evaluation benchmark in the financial domain, covering 10 major financial certifications, consisting of CPA (Certified Public Accountant), CCBP (Certification of China Banking Professional), FundPQ (Fund Practitioner Qualification), SPQ (Securities Practitioner Qualification), CICE, Economist, FuturesPQ (Futures Practitioner Qualification), CTA (Certified Tax Agents), CAA (China Actuarial Association) and AFP (Associate Financial Planner), leading to a total of 7,173 multiple-choice questions. 

FLAME~\footnote{\url{https://github.com/FLAME-ruc/FLAME/tree/main}} is a newly proposed Chinese financial benchmark, including the Financial Qualification Certification Evaluation System (FLAME-Cer) and the Financial Scenario Application Evaluation System (FLAME-Sce). 
    
    \begin{itemize}
        \item FLAME-Cer consists of 14 financial certifications: AFP (Associate Financial Planner), CAA (China Actuarial Association), CFA (Chartered Financial Analyst), CIA (Certified Internal Auditor), CISA (Certified Information Systems Auditor), CMA (Certified Management Accountant), CPA (Certified Public Accountant), FRM (Financial Risk Manager), CLIQ (Chinese Life Insurance Qualification, 8 qualification certificates), FundPQ (Fund Practitioner Qualification), FuturesPQ (Futures Practitioner Qualification), Preliminary \& Intermediate Economist, SPQ (Securities Practitioner Qualification), CCBP (Certification of China Banking Professional). 
        
        \item FLAME-Sce encompasses 10 financial scenario applications, including Financial Knowledge and Theory, Financial Compliance, Financial Document Generation, Financial Intelligent Customer Services, Financial Risk Control, Financial Document Processing, Financial Analysis and Research, Financial Investment and Wealth Management, Financial Marketing, and Financial Data Processing.
    \end{itemize}

\subsection{Preliminary Experiments}

\subsubsection{Ablation Study for Domain Self-constraint Continual Pre-training}

\begin{figure}
 \centering
\includegraphics[width=9cm]{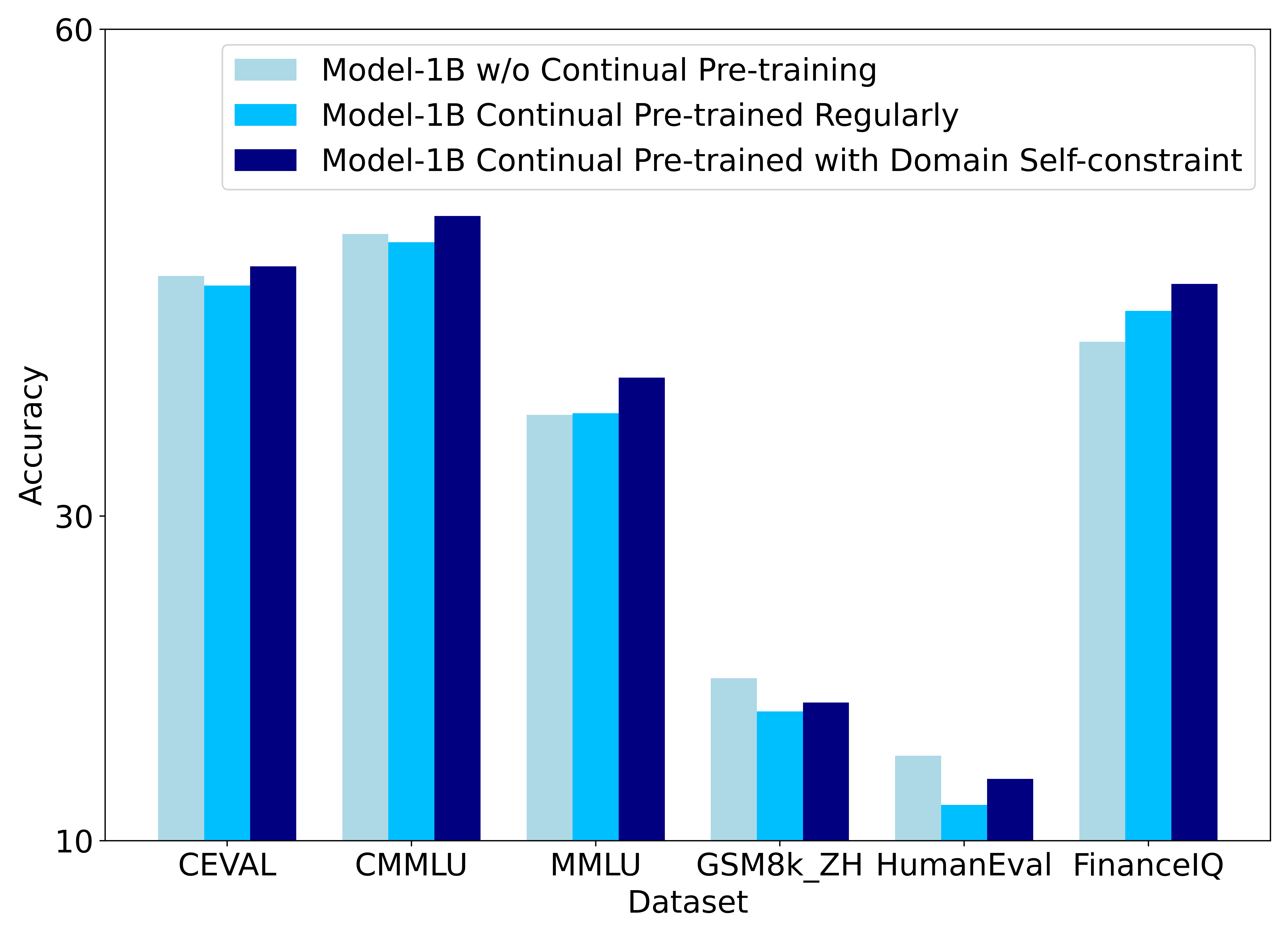}
\caption{Here we put a case on a 1B model to explore the validation of the proposed domain self-constraint training strategy on public general datasets and public financial benchmark FinanceIQ.}
\label{fig:plot_bar_part1}
\end{figure}

To evaluate the effectiveness of the domain self-constraint strategy, we conducted a preliminary experiment using a 1B small Base model, with the results presented in Figure~\ref{fig:plot_bar_part1}. The figure compares the model's performance on multiple datasets using three different training approaches: without continual pre-training, with regular continual pre-training, and with the proposed domain self-constraint continual pre-training. The results indicate that regular continual pre-training leads to a loss of general knowledge as the model learns financial knowledge. In contrast, the proposed domain self-constraint strategy enhances the model's financial performance while preserving its general knowledge.

\begin{figure}
 \centering
\includegraphics[width=9cm]{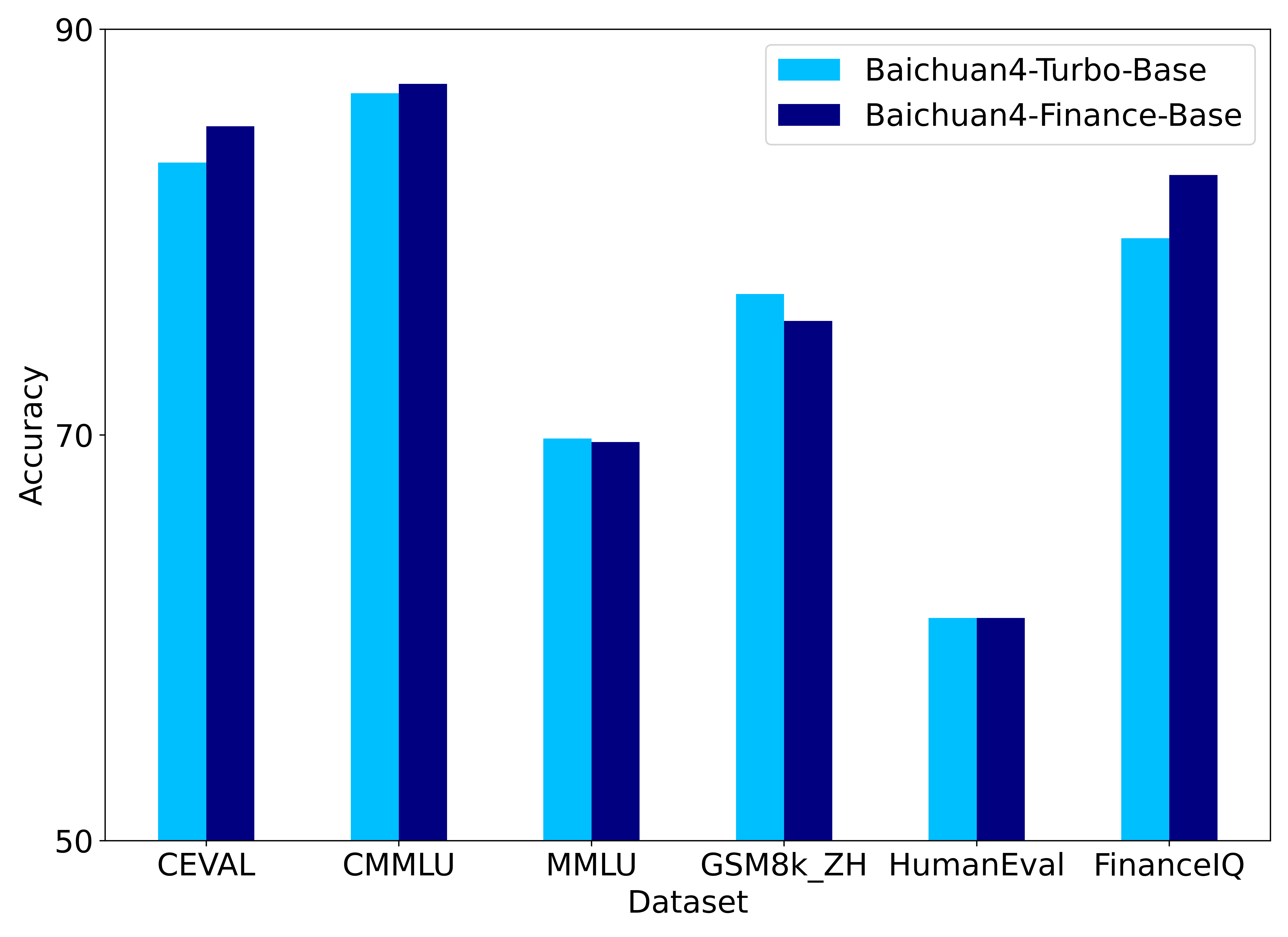}
\caption{Performance comparison across Baichuan4-Finance-Base and its backbone Baichuan4-Turbo base model on public general datasets, and financial benchmark FinanceIQ.}
\label{fig:plot_bar_part2}
\end{figure}

Furthermore, we compared the performance of Baichuan4-Finance-Base with its backbone, Baichuan4-Turbo base model, to showcase the capabilities of Baichuan4-Finance-Base on both general tasks and financial-related tasks. The results are shown in Figure~\ref{fig:plot_bar_part2}, from which we can see that Baichuan4-Finance-Base surpasses
Baichuan4-Turbo base model on financial benchmark by significant margins with comparable performance on general datasets. 

\subsubsection{Ablation Study for PPO}
\label{sec:ablation_PPO}

To investigate the necessity of the reinforcement learning phase, i.e. the effectiveness of PPO, we compared the following three models:

\begin{itemize}
    \item Baichuan4-Finance: trained Baichuan4-Finance-Base with supervised fine-tuning strategy on the SFT demonstration dataset and then aligned it by PPO strategy on the preference dataset;
     \item Baichuan4-Finance w/o PPO: trained Baichuan4-Finance-Base with supervised fine-tuning strategy, on the SFT demonstration dataset and the prompts in preference dataset with corresponding ground truth labels;
    \item Baichuan4-Finance-Base-SFT: trained Baichuan4-Finance-Base with the supervised fine-tuning strategy only on the SFT demonstration dataset.
   
\end{itemize}

\begin{figure}[ht]
 \centering
 \begin{tikzpicture}
  \node[anchor=center] (img) {\includegraphics[width=8cm]{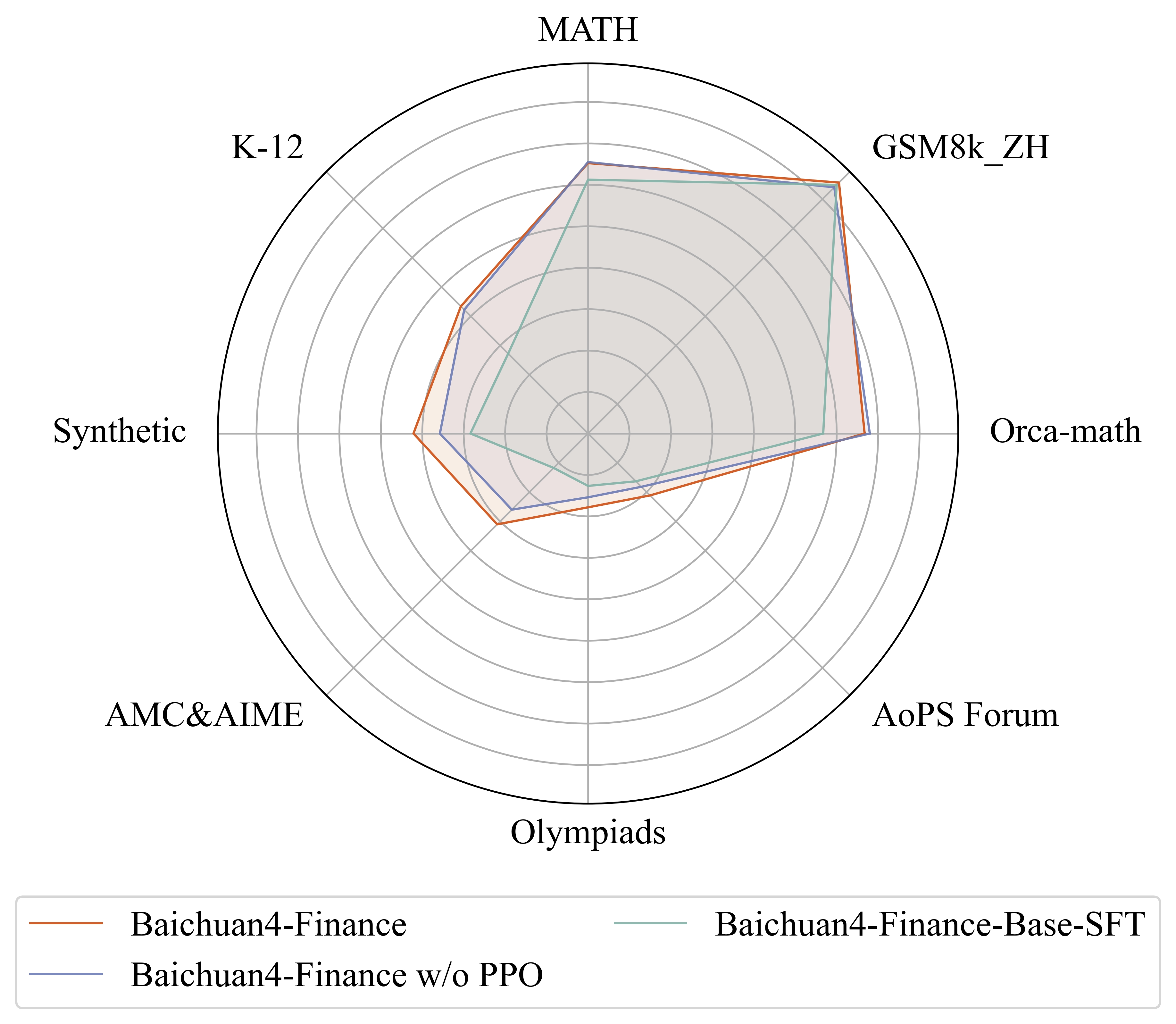}};
  \node[anchor=center, text=black] at (img)[yshift=0.5cm] {0};
    \node[anchor=center, text=black] at (img.south) [yshift=1.75cm] {0.9};
 \end{tikzpicture}
\caption{An experiment to demonstrate the effectiveness of PPO focusing on models' mathematical abilities conducted on eight mathematical datasets. }
\label{fig:ablation_PPO}
\end{figure}

The results are visualized in Figure~\ref{fig:ablation_PPO}, we could find that under the same training data, using the SFT and PPO algorithm for training performs significantly better than using the SFT algorithm alone, thereby validating the effectiveness of PPO.

Furthermore, we explored the underlying mechanism of PPO's effectiveness by conducting the following experiments:
\begin{itemize}
    \item Baichuan4-Finance pass@1: we perform a zero-shot inference with Baichuan4-Finance.
    \item Baichuan4-Finance w/o PPO pass@1: we perform a single inference with Baichuan4-Finance w/o PPO (see Section~\ref{sec:ablation_PPO} for the definition of this model);
    \item Baichuan4-Finance w/o PPO pass@5: we perform inference process for 5 runs with Baichuan4-Finance w/o PPO. The model is considered to have inferred correctly on this sample as long as at least one of the answers is correct.
\end{itemize}

\begin{figure}[ht]
 \centering
 \begin{tikzpicture}
  \node[anchor=center] (img) {\includegraphics[width=8.8cm]{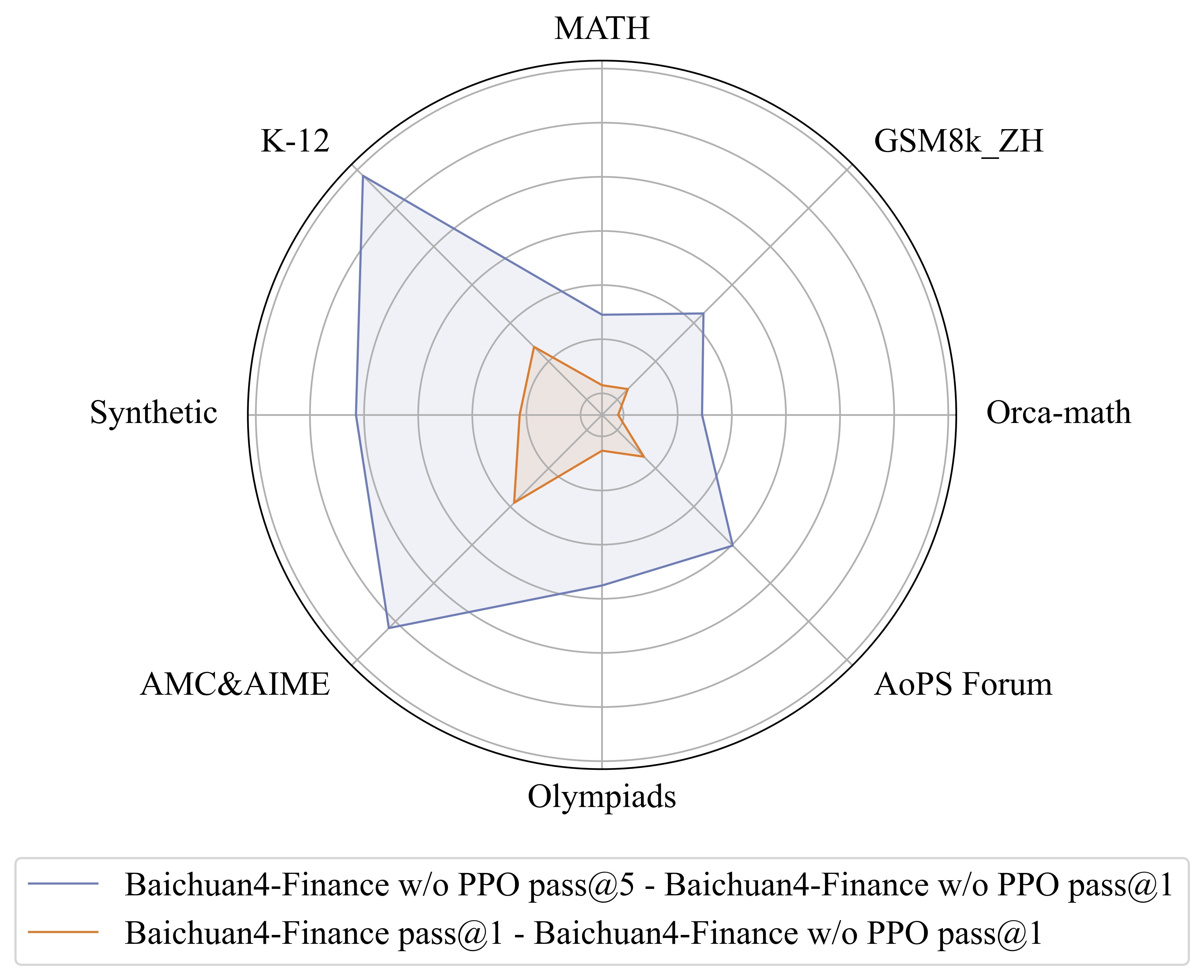}};
  \node[anchor=center, text=black] at (img)[yshift=0.6cm] {0};
    \node[anchor=center, text=black] at (img.south) [yshift=1.96cm] {0.3};
 \end{tikzpicture}
\caption{An experiment to explore the mechanism of how PPO takes effect.}
\label{fig:ablation_PPO1}
\end{figure}

It is a well-established fact that the metrics calculated using the Baichuan4-Finance w/o PPO pass@5 evaluation method are higher than those calculated using the Baichuan4-Finance w/o PPO pass@1 evaluation method. If there is a noticeable difference between these two evaluation methods, it indicates that the model's hit rate per inference is unstable. If the results of these two methods are near, PPO's potential to enhance the model is limited for its learning from the difference value between answers. Based on these analyses, we calculate the minus of Baichuan4-Finance w/o PPO pass@1's performance and Baichuan4-Finance w/o PPO pass@5's performance on eight datasets to explore the potential effectiveness of PPO. The results are shown in the blue line in the radar chart~\ref{fig:ablation_PPO1}. Besides, for comparison, we also calculate the minus of Baichuan4-Finance pass@1's performance and Baichuan4-Finance w/o PPO pass@1's performance, and the results are shown in the orange line in Figure~\ref{fig:ablation_PPO1}. By comparing these two lines, their similar shape indicates that the performance of Baichuan4-Finance pass@1 and its of Baichuan4-Finance w/o PPO pass@5 are positively correlated, which further presents that PPO could enhance the model's hit rate per inference by transferring the potential performance from pass@5.

\subsection{Main Results of Baichuan4-Finance}

\subsubsection{Baselines}
Under zero-shot setting, we compare the generation accuracy of Baichuan4-Finance with GPT-4o~\footnote{\url{https://openai.com/index/hello-gpt-4o/}} and two open-source LLMs Qwen2.5-72B-Instruct~\footnote{\url{https://huggingface.co/Qwen/Qwen2.5-72B-Instruct}} and XuanYuan3-70B~\footnote{\url{https://huggingface.co/Duxiaoman-DI/XuanYuan-70B}} on benchmark FinanceIQ. For the benchmark FLAME, we present the evaluation results of the official institute, which contains 5 competitive chat LLMs including GPT-4o, ERNIE-4.0-Turbo-128K~\footnote{\url{https://cloud.baidu.com/}}, GLM-4-PLUS~\footnote{\url{https://open.bigmodel.cn/}}, Qwen2.5-72B-Instruct and XuanYuan3-70B.

\subsubsection{Comparison Results}

The comparison results of Baichuan4-Finance and baselines on the FinanceIQ dataset are shown in Table~\ref{tab:performance_financeIQ}, while the results of the FLAME benchmark are shown in Table~\ref{tab:performance_flame_cer} and Table~\ref{tab:performance_flame_sce}. We can see that Baichuan4-Finance has strong capabilities in tackling different financial certifications and application scenarios.

\begin{table}[ht]
\centering
\caption{The zero-shot accuracy of Baichuan4-Finance and baselines on FLAME-Cer benchmark.}
\resizebox{\linewidth}{!}{\begin{tabular}{ccccccc}
    \toprule
    \textbf{FLAME-Cer} & \textbf{GPT-4o} & \textbf{\makecell{ERNIE-4.0-\\Turbo-128K}} & \textbf{\makecell{GLM-4-\\PLUS}} & \textbf{\makecell{Qwen2.5-72B-\\Instruct}} & \textbf{\makecell{XuanYuan3-70B-\\Chat}} & \textbf{Baichuan4-Finance} \\
    \midrule
    \textbf{AFP} & 74.19 & 72.93 & 79.95 & 79.37 & 64.09 & \textbf{90.08} \\
  \textbf{CFA} & \textbf{86.97} & 74.44 & 81.45 & 82.92 & 63.88 & 85.59 \\
   \textbf{CAA} & 44.65 & 48.37 & 43.26 & 55.81 & 31.63 & \textbf{66.51} \\
   \textbf{CIA} & 84.46 & 77.94 & 83.46 & 86.69 & 70.53 & \textbf{92.59} \\
  \textbf{CISA} & 86.72 & 77.94 & 83.96 & 86.83 & 70.31 & \textbf{95.76} \\
  \textbf{CMA} & 83.21 & 73.68 & 76.69 & \textbf{86.10} & 65.71 & \textbf{86.10} \\
   \textbf{CPA} & 68.67 & 70.18 & 78.95 & 85.31 & 68.64 & \textbf{93.16} \\
  \textbf{FRM} & 74.94 & 67.92 & 76.69 & 76.87 & 55.46 & \textbf{82.87} \\
  \textbf{CLIQ} & 81.20  & 79.95 & 81.70  & 86.61 & 69.10  & \textbf{93.82} \\
  \textbf{FundPQ} & 82.96 & 84.21 & 84.71 & 94.61 & 76.73 & \textbf{97.93} \\
  \textbf{FuturesPQ} & 75.94 & 78.45 & 82.46 & 90.23 & 73.74 & \textbf{96.54} \\
  \textbf{Ecomonist} & 80.45 & 87.97 & 90.48 & 93.41 & 82.35 & \textbf{95.78} \\
   \textbf{SPQ} & 76.69 & 84.21 & 88.72 & 94.33 & 78.74 & \textbf{97.60} \\
  \textbf{CCBP} & 78.70  & 86.97 & 86.47 & 92.72 & 79.26 & \textbf{96.42} \\
\midrule   \textbf{Average} & 78.23 & 77.03 & 81.17 & 88.24 & 72.07 & \textbf{93.62} \\
    \bottomrule
    \end{tabular}}%
\label{tab:performance_flame_cer}
\end{table}

\begin{table}[ht]
\centering
\caption{The performance of Baichuan4-Finance and baselines on FLAME-Sce benchmark.}
\resizebox{\linewidth}{!}{
    \begin{tabular}{ccccccc}
    \toprule
    \textbf{FLAME-Sce} & \textbf{GPT-4o} & \textbf{\makecell{ERNIE-4.0-\\
    Turbo-128K}} & \textbf{\makecell{GLM-4-\\PLUS}} & \textbf{\makecell{Qwen2.5-72B-\\Instruct}} & \textbf{\makecell{XuanYuan3-70B-\\Chat}} & \textbf{Baichuan4-Finance} \\
    \midrule
    \textbf{Knowledge and Theory } & 89.14 & 87.67 & 87.06 & 87.75 & 76.94 & \textbf{91.17} \\
 \midrule
    \textbf{Compliance} & 80.27 & 80.27 & 86.95 & 83.61 & 40.13 & \textbf{87.24} \\
 \midrule
    \textbf{Document Generation} & 60.01 & 58.89 & 61.11 & 58.89 & 41.11 & \textbf{70.03} \\
 \midrule
    \textbf{\makecell{Intelligent Consumer\\ Services}} & 83.52 & 79.59 & 79.78 & 83.18 & 62.37 & \textbf{86.92} \\
 \midrule
    \textbf{Risk Control} & 82.94 & 75.86 & 80.3  & 81.99 & 63.93 & \textbf{85.36} \\
 \midrule
    \textbf{Document Processing} & 83.51 & 80.22 & 65.88 & 80.18 & 65.91 & \textbf{86.77} \\
 \midrule
    \textbf{Data Processing} & 86.43 & 89.99 & 79.87 & 91.61 & 71.92 & \textbf{91.71} \\
 \midrule
    \textbf{Analysis and Research} & 45.45 & \textbf{48.48} & 42.42 & 42.42 & 33.33 & 45.45 \\
 \midrule
    \textbf{Marketing} & 57.58 & \textbf{93.94} & 75.76 & 54.55 & 30.3  & 78.79 \\
 \midrule
    \textbf{\makecell{Investment \& \\ Wealth Management}} & 85.71 & 71.43 & \textbf{89.29} & 85.71 & 67.86 & \textbf{89.29} \\
    \midrule
    \textbf{Average} & 79.88 & 78.01 & 76.92 & 79.18 & 61.34 & \textbf{84.15} \\
    \bottomrule
    \end{tabular}%
}
\label{tab:performance_flame_sce}
\end{table}

\begin{table}[ht]
\centering
\caption{The zero-shot accuracy of Baichuan4-Finance and baselines on FinanceIQ benchmark.}
\resizebox{0.9\linewidth}{!}{
    \begin{tabular}{ccccc}
    \toprule
    \textbf{FinanceIQ} & \textbf{GPT-4o} & \textbf{Qwen2.5-72B-Instruct} & \textbf{XuanYuan3-70B-Chat} & \textbf{Baichuan4-Finance} \\
    \midrule
    \textbf{CPA} & 62.93 & 74.83 & 65.45 & \textbf{80.70} \\

    \textbf{CCBP} & 74.28 & 85.85 & 73.84 & \textbf{88.26} \\

    \textbf{SPQ} & 69.13 & 79.25 & 71.09 & \textbf{82.40} \\

    \textbf{FindPQ} & 72.02 & 83.14 & 73.97 & \textbf{83.49} \\

    \textbf{CICE} & 73.71 & 80.03 & 59.77 & \textbf{82.47} \\

    \textbf{Economist} & 82.12 & 90.19 & 67.31 & \textbf{93.85} \\

    \textbf{CTA} & 49.80  & 74.59 & 47.75 & \textbf{75.00} \\

    \textbf{FuturesPQ} & 69.52 & 80.60  & 73.67 & \textbf{85.91} \\

    \textbf{AFP} & 71.53 & 76.61 & 67.43 & \textbf{82.71} \\

    \textbf{CAA} & 37.50 & \textbf{45.45} & 37.50 & 37.50 \\
    \midrule
    \textbf{Average} & 66.25 & 77.06 & 63.78 & \textbf{79.23} \\
    \bottomrule
    \end{tabular}%
    }
\label{tab:performance_financeIQ}
\end{table}

\clearpage

\section{Conclusion}
\label{sec:conclusion}

In this report, we introduce the Baichuan4-Finance series, which includes two models: Baichuan4-Finance-Base and the chat model Baichuan4-Finance. There are three key technical highlights in this report: (1) For the data side: the continual pre-training data construction pipeline and the data mixture ratio determining process using two scaling laws; (2) For the training side, the proposed a novel pre-training strategy, called domain self-constraint training, tailored for the domain LLM continual pre-training. We evaluate Baichuan4-Finance through extensive experiments and the results indicate that Baichuan4-Finance-Base outperforms nearly all competitive baselines on financial tasks by substantial margins, without compromising performance on general LLM benchmarks. Additionally, Baichuan4-Finance delivers even more remarkable performance in financial application scenarios, highlighting its potential to drive innovation within the financial LLM community.

\clearpage

\bibliography{main}
\bibliographystyle{iclr2023_conference}

\end{CJK}
\end{document}